\documentclass[preprint,12pt]{elsarticle}

\usepackage{booktabs}

\usepackage{amssymb}
\usepackage{amsmath}

\journal{Nuclear Physics B}

\begin{document}

\begin{frontmatter}

\title{TGC-Net: A Structure-Aware and Semantically-Aligned Framework for Text-Guided Medical Image Segmentation}

\author[1]{Gaoren Lin}
\ead{lingaoren@whu.edu.cn}

\author[1]{Huangxuan Zhao\corref{cor1}}
\ead{zhao_huangxuan@sina.com}

\author[2]{Yuan Xiong}
\ead{xiongyuan@hust.edu.cn}

\author[1]{Lefei Zhang\corref{cor1}}
\ead{zhanglefei@whu.edu.cn}

\author[1]{Bo Du\corref{cor1}}
\ead{dubo@whu.edu.cn}

\author[2]{Wentao Zhu\corref{cor1}}
\ead{tjgkzwt@163.com}

\cortext[cor1]{Corresponding author.}

\affiliation[1]{organization={School of Computer Science, Wuhan University},
            city={Wuhan},
            postcode={430072},
            state={Hubei},
            country={China}}

\affiliation[2]{organization={Department of Orthopedics, Tongji Hospital,
            Tongji Medical College, Huazhong University of Science and Technology},
            city={Wuhan},
            postcode={430030},
            state={Hubei},
            country={China}}

\begin{abstract}
Text-guided medical segmentation enhances segmentation accuracy by utilizing clinical reports as auxiliary information. However, existing methods typically rely on unaligned image and text encoders, which necessitate complex interaction modules for multimodal fusion. While CLIP provides a pre-aligned multimodal feature space, its direct application to medical imaging is limited by three main issues: insufficient preservation of fine-grained anatomical structures, inadequate modeling of complex clinical descriptions, and domain-specific semantic misalignment. To tackle these challenges, we propose TGC-Net, a CLIP-based framework focusing on parameter-efficient, task-specific adaptations. Specifically, it incorporates a Semantic–Structural Synergy Encoder (SSE) that augments CLIP’s ViT with a CNN branch for multi-scale structural refinement, a Domain–Augmented Text Encoder (DATE) that injects large-language-model–derived medical knowledge, and a Vision–Language Calibration Module (VLCM) that refines cross-modal correspondence in a unified feature space. Experiments on five datasets across chest X-ray and thoracic CT modalities demonstrate that TGC-Net achieves state-of-the-art performance with substantially fewer trainable parameters, including notable Dice gains on challenging benchmarks.
\end{abstract}

\begin{graphicalabstract}
\end{graphicalabstract}

\begin{highlights}
\item We propose TGC-Net, CLIP-based framework for text-guided medical segmentation.
\item Semantic–Structural Encoder fuses CLIP ViT and CNN for structure-aware visuals.
\item Domain–Augmented Text Encoder injects LLM-based medical knowledge into CLIP.
\item Lightweight Calibration Module refines image–text alignment with few params.
\item TGC-Net achieves SOTA Dice and mIoU on 5 datasets with just 10M trainable params.
\end{highlights}


\end{frontmatter}



\section{Introduction}
\label{sec1}
\begin{figure}[t]
    \centering
    \includegraphics[width=0.95\linewidth]{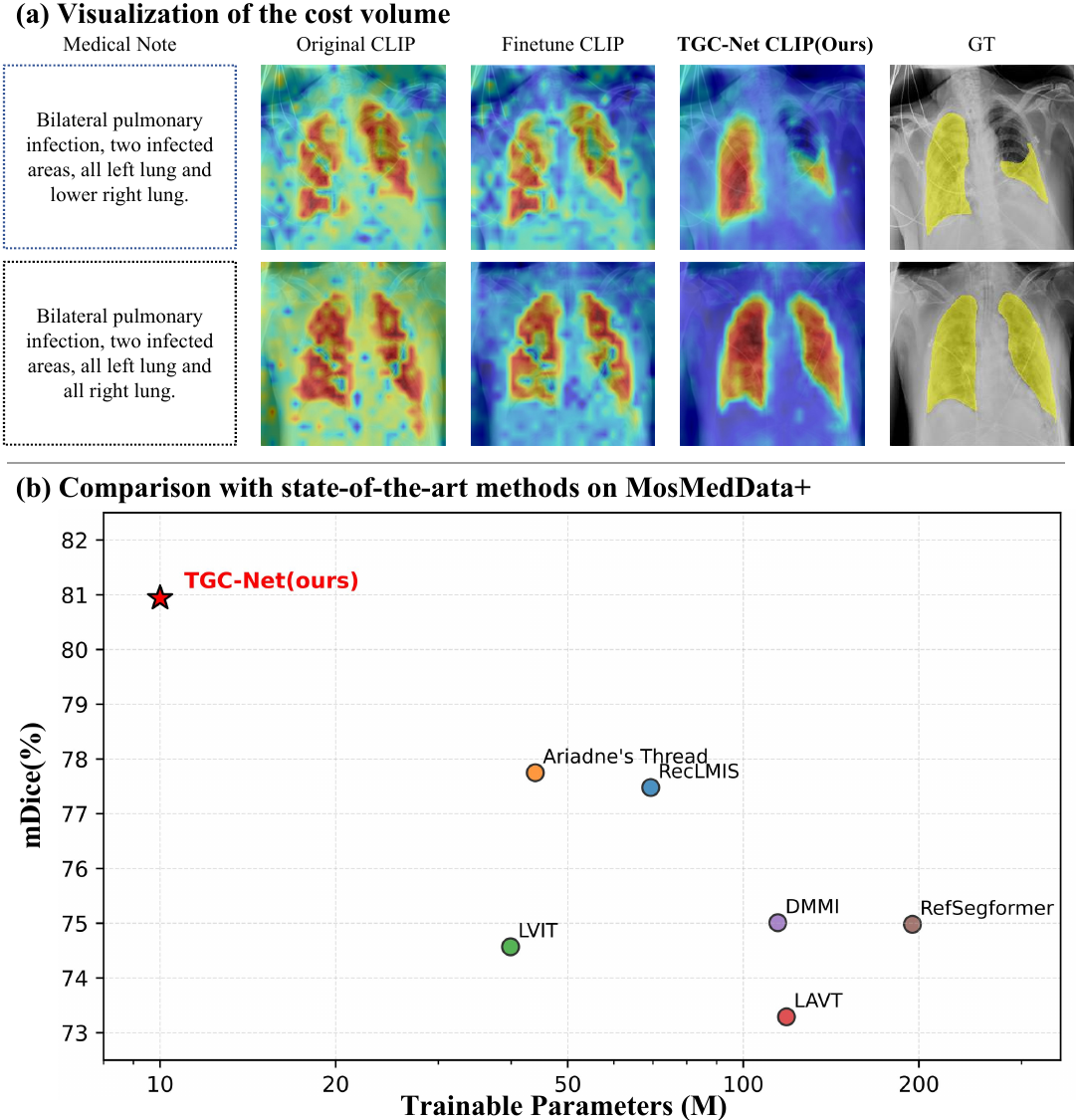}
    \caption{
    TGC-Net: Achieving precise localization via parameter-efficient adaptation.
    (a) Qualitatively, we visualize the feature similarity maps between text prompts and image patches. Standard and fine-tuned CLIP suffer from severe noise and misalignment (columns 2-3), whereas our method produces a purified map that precisely highlights the infection area.
    (b) Quantitatively, benefiting from our proposed task-specific modules, TGC-Net achieves state-of-the-art accuracy (mDice) on MosMedData+ with substantially fewer trainable parameters than existing methods.
    }
    \label{fig:motivation}
\end{figure}

\begin{figure}[t]
    \centering
    \includegraphics[width=1\linewidth]{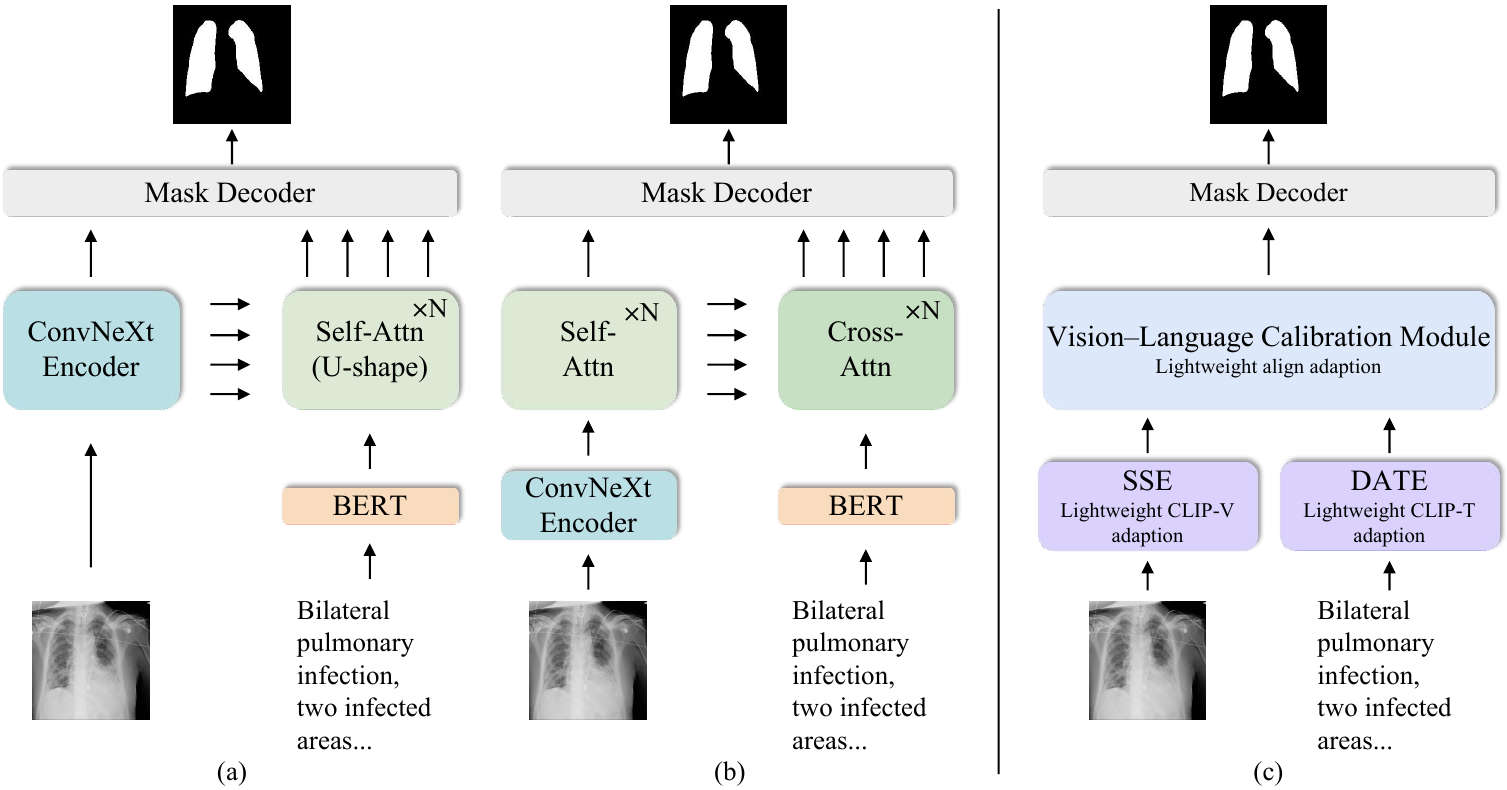}
    \caption{
    Overall architectural comparison. (a) Typical text-guided medical segmentation frameworks rely on heavy dual encoders with stacked cross-modal interaction and U-shaped decoders. (b) The dual-branch fusion architecture employs N layers of cross-attention in the text branch to align visual and textual features. (c) Our proposed TGC-Net builds on pre-aligned CLIP encoders and introduces lightweight SSE, DATE, and VLCM modules, resulting in a more compact yet expressive architecture.
    }
    \label{fig:architecture_compare}
\end{figure}

Medical image segmentation (MIS) serves as a cornerstone for computer-aided diagnosis, treatment planning, and clinical decision support~\cite{azad2022medical,huang2025multi, liu2025cswin}. Despite remarkable progress in deep learning, most segmentation models still rely heavily on large-scale, pixel-level annotated datasets. In the medical domain, collecting such annotations is labor-intensive, requires expert knowledge, and incurs substantial costs, which severely limits scalability and generalization~\cite{huang2024cross, li2023lvit}.

In contrast, medical reports are routinely acquired alongside imaging studies and naturally encode rich semantic and diagnostic information~\cite{li2023lvit}. These texts contain expert prior knowledge about lesion type, location, and context, providing a valuable yet underutilized supervision signal. Leveraging such complementary cues has motivated the task of language-guided medical image segmentation, which aims to align textual descriptions with corresponding image regions for more interpretable and more accurate segmentation~\cite{zhong2023ariadne, feng2024enhancing, wang2025towards}.

Existing approaches to text-guided medical segmentation typically adopt a dual-encoder paradigm: a specialized visual backbone (e.g., ConvNeXt) and a separate text encoder (e.g., BioBERT) are jointly trained, and their features are fused through multiple layers of cross-modal interaction\cite{li2023lvit}\cite{zhong2023ariadne}\cite{wang2025towards}. As illustrated in Fig.~\ref{fig:architecture_compare}(a)–(b), such frameworks often rely on stacked cross-attention blocks, multi-stage fusion pyramids, and heavy U-shaped decoders to learn task-specific alignment between image and text representations. While effective, these designs introduce a large number of trainable parameters that must be optimized on relatively small datasets, making them prone to overfitting. In essence, current methods compensate for the lack of pre-existing vision–language alignment by building increasingly complex interaction modules, at the cost of efficiency and robustness.

Meanwhile, the advent of large-scale vision–language models (VLMs) such as CLIP has fundamentally changed the landscape of multimodal learning. Trained on hundreds of millions of image–text pairs, CLIP jointly optimizes an image encoder and a text encoder to inhabit a shared semantic space, where global visual features are already well aligned with their textual descriptions\cite{radford2021clip}\cite{cho2024cat-seg}\cite{zhang2025perceive}. This raises an appealing question for medical applications: instead of designing heavy cross-modal interaction modules from scratch, can we reuse the pre-aligned CLIP space and perform only lightweight, task-specific adaptation for text-guided referring segmentation?

A straightforward approach is to directly apply the frozen CLIP encoders to medical images and corresponding reports. As shown in Fig.~\ref{fig:motivation}(a), by computing the similarity between text features and image patch features, the raw CLIP visual representations exhibit a coarse localization capability—for pulmonary-related prompts, high-response regions tend to concentrate around the lung areas. However, these activations remain scattered, noisy, and poorly aligned with lesion boundaries, resulting in unsatisfactory segmentation masks. Simply fine-tuning a small subset of CLIP parameters (e.g., the query and value projections in attention layers) can sharpen the response focus, as illustrated in Fig. 1(a), yet the predicted masks still suffer from inaccurate shapes and residual activations in irrelevant regions. These observations indicate that directly transplanting CLIP to text-guided medical image segmentation is insufficient, and more targeted adaptations are necessary to bridge the gap between general vision-language pretraining and fine-grained, pixel-level reasoning.

A deeper analysis reveals three main challenges that hinder the direct adaptation of CLIP to this task. First, a structural gap exists on the visual side: while CLIP’s vision encoder, usually a Vision Transformer (ViT), excels at capturing global semantics, it is less effective at preserving fine-grained structural details, such as organ boundaries or subtle nodules, which are critical for precise segmentation. Second, a semantic gap persists on the textual side: CLIP is pretrained on natural images paired with concise, everyday captions, leaving it under-equipped to represent complex medical narratives with domain-specific terminology and implicit clinical context. Third, a calibration gap affects cross-modal alignment: although CLIP embeds both visual and textual features into a shared space, this alignment is learned on natural image-text distributions and may exhibit systematic biases when transferred to medical domains, particularly for subtle or rare pathologies.

In this work, we propose the \textbf{T}ri‑\textbf{G}ap \textbf{C}alibration \textbf{Net}work (TGC‑Net), a CLIP-based framework explicitly designed to address these three gaps and adapt pre-aligned vision-language models to text-guided medical segmentation. Our core idea is to preserve CLIP’s global alignment strength while introducing targeted, lightweight enhancements for structure, semantics, and cross-modal calibration. To mitigate the structural gap, we design a \textbf{S}emantic–\textbf{S}tructural Synergy \textbf{E}ncoder (SSE), which augments the CLIP visual encoder with a lightweight CNN branch. This branch extracts multi-scale, fine-grained structural features, which are then fused with the global semantic tokens from the ViT through a guided fusion module. The resulting visual representations retain semantic alignment with text while being significantly more informative for pixel-level segmentation. To address the semantic gap, we introduce a \textbf{D}omain–\textbf{A}ugmented \textbf{T}ext \textbf{E}ncoder (DATE): starting from the original clinical report, we leverage a large language model to generate auxiliary medical knowledge, which is then injected into the CLIP text embeddings, yielding medically enriched textual representations that better correspond to visual disease patterns. Finally, to correct the calibration gap, we propose a \textbf{V}ision–\textbf{L}anguage \textbf{C}alibration \textbf{M}odule (VLCM), which operates in a unified feature space and employs a lightweight interaction mechanism to refine the correspondence between textual tokens and visual patch features. The overall architecture, shown in Fig.~\ref{fig:architecture_compare}(c), remains notably more compact than conventional frameworks that learn cross-modal alignment entirely from scratch.

Extensive experiments are conducted on five widely used benchmark datasets and the results prove the superiority of our TGC-Net.The main contributions of this work are summarized as follows:
\begin{itemize}
    \item We present TGC-Net, a CLIP-based framework for text-guided medical image segmentation that systematically leverages the pre-aligned vision–language space of CLIP while introducing targeted adaptations for medical images and reports, explicitly addressing structural, semantic, and calibration gaps.
    \item We propose the SSE and the DATE to enhance visual and textual representations. the SSE combines a lightweight CNN branch with the CLIP ViT to deliver semantically aligned yet structurally precise visual features; and the DATE enriches the text embedding space by injecting medical knowledge derived from large language models, resulting in robust and clinically informed text representations.
    \item We design the VLCM, a lightweight alignment mechanism that operates in a unified feature space and effectively refines the cross-modal correspondence between medical images and textual descriptions. 
    \item Extensive experiments and ablation studies on five datasets spanning X-ray and CT modalities demonstrate that TGC‑Net achieves new state-of-the-art performance in text-guided medical image segmentation.
\end{itemize}

\section{Related Work}
\subsection{Medical Image Segmentation}
Medical Image Segmentation (MIS) is a long-standing and fundamental task in medical image analysis\cite{munia2025attention}. The field was initially dominated by deep learning, particularly Convolutional Neural Networks (CNNs)\cite{du2020medical}\cite{jha2020doubleu}\cite{kayalibay2017cnn}. The U-Net architecture \cite{ronneberger2015u}, with its iconic encoder-decoder structure and skip-connections, set the gold standard by demonstrating an exceptional ability to capture fine-grained local details and produce precise boundary maps. Its variants and subsequent improvements, such as nnU-Net \cite{isensee2021nnu}, have shown remarkable performance and robustness across a wide array of medical imaging modalities.

Despite their success, the limited receptive field of CNNs restricts their ability to model long-range contextual relationships. To address this, Transformers \cite{vaswani2017attention} were introduced, leading to two main integration strategies. First, hybrid architectures like TransUNet \cite{chen2021transunet} emerged, which strategically embed a Transformer block into the bottleneck of a CNN-based U-Net to capture global context, while still relying on the convolutional path for spatial feature extraction and upsampling. Second, pure Transformer backbones, such as in Swin-UNETR \cite{hatamizadeh2021swin}, were proposed to replace the entire CNN encoder, using a hierarchical Transformer (Swin Transformer) as the primary feature extractor.

This evolution highlights a critical trade-off: while Transformers provide unparalleled global context modeling, they often lack the strong inductive biases of CNNs for extracting high-resolution, fine-grained local features, which are critical for precise medical boundary delineation\cite{li2023lvit}\cite{xia2025comprehensive}. This inherent architectural compromise, which sacrifices local precision for global context, is what we identify as the Structural Gap.Our work addresses this gap through the SSE, a hybrid design that preserves the powerful global semantic representations produced by the CLIP visual encoder while enriching them with fine-grained, segmentation-specific structural cues from a complementary CNN branch. By synergistically fusing these two sources of information, SSE produces robust, structurally aware visual features tailored for text-guided medical segmentation.

\subsection{CLIP and Medical Domain Adaptation}
The emergence of large-scale Vision-Language Models (VLMs), pre-trained on hundreds of millions of image-text pairs, has revolutionized multi-modal research. CLIP \cite{radford2021clip} is a canonical example, learning a unified embedding space where corresponding image and text features are projected to be close. This inherent alignment provides a powerful foundation, eliminating the need for manually designing complex fusion modules, a limitation seen in earlier works like LViT \cite{li2023lvit}. LViT, while pioneering the language-guided medical task, employed separate and unaligned BERT and CNN encoders, thus failing to leverage this powerful pre-alignment.

Despite their potential, directly applying general-domain VLMs like CLIP to medicine is highly challenging due to a significant domain shift. This shift manifests in several ways that are central to our work. First, beyond the aforementioned Structural Gap in the ViT backbone, a critical Semantic Gap exists, as the training vocabulary of CLIP is based on general-domain text and lacks the understanding of complex, specialized medical terminology (e.g., ``hypoechoic," ``parenchymal heterogeneity"). Second, and most critically, the core issue is that the alignment learned by CLIP (e.g., matching a photo of a ``dog" to the word ``dog") does not transfer to the medical domain (e.g., matching a ``hypodense lesion" on a CT scan to its textual description). This constitutes the Alignment Gap that we identify.

Several recent works have focused on adapting VLMs to medicine to solve these issues. For example, GLoRIA \cite{huang2021gloria} and MedCLIP \cite{wang2022medclip} use contrastive learning on medical-specific image-text pairs. More specific to our task, CAT \cite{huang2024cat} has attempted to address the Semantic Gap by enriching textual prompts with medical domain knowledge. Most relevantly, RecLMIS \cite{huang2024cross} proposes a cross-modal reconstruction task to explicitly re-establish the relationship between modalities, directly tackling the Alignment Gap.While these works have begun to address parts of the problem (e.g., semantics in CAT, alignment in RecLMIS), a systematic framework that simultaneously accounts for structure, semantics, and alignment is still necessary. Our proposed framework, with its SSE, DATE, and VLCM modules, is designed to be this comprehensive solution.

\section{Method}

We propose TGC-Net, an end-to-end framework built upon CLIP for language-guided medical image segmentation. As discussed in the introduction, the direct application of VLMs faces three core challenges: a Structural Gap, a Semantic Gap, and an Alignment Gap. To address these in a systematic manner, TGC-Net incorporates three dedicated modules, each designed to bridge one of these gaps.

\begin{figure*}[t]
    \centering
    \includegraphics[width=0.95\linewidth]{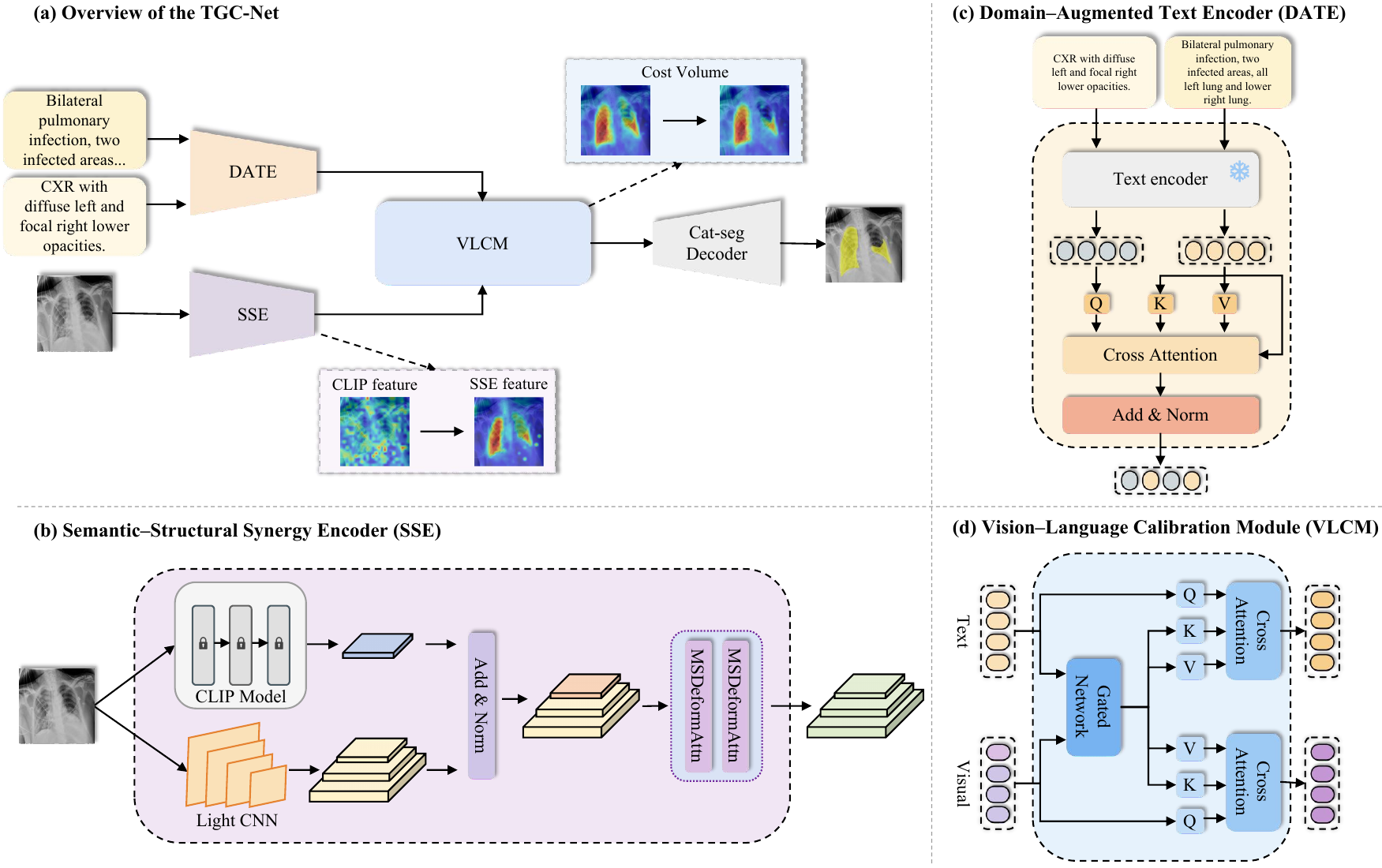 }
    \caption{(a) Overview of the proposed Tri-Gap Calibration Network (TGC-Net). The framework comprises three core components that jointly address the Structural, Semantic, and Alignment gaps in language-guided medical image segmentation. (b) The SSE integrates global semantic features from the CLIP visual encoder with fine-grained structural cues extracted by a lightweight CNN, producing a feature representation enriched with both high-level semantic information and fine-grained structural details. (c) The DATE enhances the primary text prompt with domain-specific medical semantics via cross-attention, generating knowledge-infused textual features. (d) The VLCM refines and re-aligns visual and textual embeddings within the medical domain through gated cross-attention interactions. The calibrated multi-modal features are finally fed into the CAT-Seg decoder to generate the segmentation mask. }
    \label{fig:model_pdf}
\end{figure*}

As illustrated in Fig.\ref{fig:model_pdf} (a), the framework processes the input image $I$ through the SSE, which fuses local structural features from a CNN with global semantic information from a Vision Transformer (ViT) to resolve the Structural Gap. Concurrently, the input text prompts are processed by the DATE, which injects domain-specific medical knowledge into the text representations, thus bridging the Semantic Gap.

The visual features $F_V$ from the SSE and the textual features $F_T$ from the DATE are then passed into the VLCM, which re-aligns the features in the medical domain, resolving the \textit{Alignment Gap}.

Finally, the calibrated features $F'_V$ and $F'_T$ are passed to the Cost-Aggregating Decoder, inspired by CAT-Seg, to compute a similarity cost volume and aggregate the information to produce the final, precise segmentation mask $M$.

The following subsections provide a detailed description of each module's design, implementation, and how each contributes to bridging the three critical gaps in the domain of medical image segmentation.

\subsection{Semantic–Structural Synergy Encoder}

The Semantic–Structural Synergy Encoder (SSE) is designed to bridge the \textit{semantic–structural gap} inherent in CLIP. Although the CLIP visual encoder, typically implemented as a Vision Transformer (ViT), is highly effective at capturing global contextual semantics, it lacks sensitivity to fine-grained structural cues and boundary details that are critical for precise medical segmentation. To address this limitation, SSE establishes a synergistic interaction between CLIP’s global semantic representations and localized structural priors extracted by a CNN branch, achieving both semantic coherence and spatial precision in the visual feature space.

As illustrated in Fig.~\ref{fig:model_pdf} (b), SSE adopts a dual-branch design that processes an input image $I \in \mathbb{R}^{H \times W \times 3}$ in parallel.  
The Global-Semantic Branch employs the pre-trained CLIP ViT encoder $\Phi_V$, whose parameters remain frozen to preserve its general visual understanding. This branch outputs a high-level semantic feature $F_{clip} \in \mathbb{R}^{N_v \times D_v}$ from the final layer.  
The Local-Structural Branch consists of a lightweight CNN encoder $\Phi_C$, which captures fine-grained spatial structures at multiple resolutions, producing a feature hierarchy $\{F_{cnn}^{1/8}, F_{cnn}^{1/16}, F_{cnn}^{1/32}\}$.

To unify the two modalities, SSE introduces a semantic–structural fusion mechanism. The global semantic feature $F_{clip}$ is projected by a linear layer $P_V$ to match the channel dimension of the deepest CNN feature $F_{cnn}^{1/32}$, which is similarly projected by $P_C$. The two are fused via element-wise addition followed by Layer Normalization:
\begin{equation}
F_{fused}^{1/32} = \text{LayerNorm}(P_V(F_{clip}) + P_C(F_{cnn}^{1/32})).
\end{equation}
This operation produces a semantically enriched deep feature that combines CLIP’s contextual abstraction with CNN-derived structural awareness.

The fused feature is then integrated with higher-resolution CNN features to form a hybrid feature pyramid:
\begin{equation}
P_{hybrid} = \{F_{cnn}^{1/8}, F_{cnn}^{1/16}, F_{fused}^{1/32}\}.
\end{equation}
This pyramid is refined using a Multi-Scale Deformable Attention (MSDeformAttn) module~\cite{zhu2020deformable}, which dynamically samples informative spatial regions and scales for adaptive alignment between global semantics and local structures. Formally,
\begin{equation}
F_V = \text{MSDeformAttn}(P_{hybrid}),
\end{equation}
where $\text{MSDeformAttn}(\cdot)$ denotes the multi-scale deformable attention operation that aggregates cross-scale features to produce a semantically and structurally coherent representation $F_V$.

Finally, SSE outputs a hierarchical visual representation, 
\begin{equation}
    F_V = \{F_V^{1/8}, F_V^{1/16}, F_V^{1/32}\}
\end{equation}
, 
where each level corresponds to a specific spatial resolution and functional role. Among them, $F_V^{1/32}$ serves as the primary semantic–structural representation and is fed into the VLCM for fine-grained alignment with textual embeddings. The calibrated feature is then used to compute the cost matrix and guide the spatial aggregation process, ensuring precise cross-modal correspondence.

Meanwhile, the higher-resolution features $F_V^{1/16}$ and $F_V^{1/8}$ preserve detailed structural information essential for segmentation boundary recovery. These are passed into the Upsampling Decoder, where hierarchical upsampling and skip connections progressively reconstruct fine-grained segmentation maps with high structural fidelity.


\subsection{Domain–Augmented Text Encoder}

Standard vision–language models (VLMs) such as CLIP are pretrained on simple, generic captions (e.g., ``a photo of a liver''), which introduces a semantic gap when applied to the medical domain. In clinical reports, text descriptions are often long, structured, and terminology-rich (e.g., ``a well-defined, round, low-density lesion in the right hepatic lobe''), making direct utilization of such text degrade alignment performance. Conversely, using overly simplified prompts discards valuable domain-specific prior knowledge.

To bridge this contradiction, we design the DATE based on the principle of condensed knowledge infusion. As illustrated in Fig.~\ref{fig:model_pdf} (c), two types of text prompts are introduced: a concise Primary Prompt ($T_p$), e.g., ``Bilateral pulmonary infection, two infected areas, all left lung and lower right lung.'' and a domain-rich Auxiliary Prompt ($T_a$), e.g., ``CXR with diffuse left and focal right lower opacities.'' The auxiliary prompt is automatically generated by a Large Language Model (LLM) to encapsulate high-level clinical semantics.

Both prompts are first encoded by the same pretrained text encoder $\Phi_T$ (e.g., CLIP Text Encoder) to obtain their respective embeddings:
\[
F_p = \Phi_T(T_p), \quad F_a = \Phi_T(T_a).
\]

Our objective is to infuse the domain knowledge from $F_a$ into $F_p$ while maintaining the efficiency and compactness of the primary prompt.

To achieve this, DATE employs a Cross-Attention mechanism, where the auxiliary feature $F_a$ acts as the \textit{Query} (Q), and the primary feature $F_p$ serves as both \textit{Key} (K) and \textit{Value} (V):
\begin{equation}
F_{\text{infused}} = \text{CrossAttn}(Q=F_a, K=F_p, V=F_p).
\end{equation}

This formulation enables the auxiliary feature to selectively extract and reinforce semantically relevant components from the primary representation.

The resulting knowledge-infused feature $F_{\text{infused}}$ is then integrated with the primary feature through a residual connection followed by layer normalization:
\begin{equation}
F_T = \text{LayerNorm}(F_p + F_{\text{infused}}).
\end{equation}

The final output $F_T \in \mathbb{R}^{L \times D_t}$ constitutes a semantically enriched yet compact text representation. It retains the clarity of the primary prompt while absorbing domain-specific knowledge from the auxiliary prompt, forming a robust textual foundation for subsequent Vision–Language Calibration and segmentation decoding.

\subsection{Vision--Language Calibration Module}
\label{subsec:vlcm}

Although pretrained vision--language models (VLMs) exhibit strong general alignment ability, their cross-modal correspondences are often unreliable in the specialized medical domain, resulting in an Alignment Gap. To address this issue, we introduce the Vision--Language Calibration Module (VLCM). As illustrated in Fig.~\ref{fig:model_pdf} (d), unlike standard cross-attention mechanisms that directly interact raw features, our VLCM employs a Generative Gating strategy. It constructs a shared, adaptive context via a Gated Network to calibrate the visual feature $F_V$ and textual feature $F_T$ simultaneously.

The calibration process is formulated as a coherent flow involving contextual integration and bidirectional refinement:

\textbf{Adaptive Context Generation.}
First, rather than interacting independently, both the visual feature $F_V$ and textual feature $F_T$ are fed into a shared Gated Network $\mathcal{G}(\cdot)$. This network aims to distill the most informative semantic cues from both modalities and suppress noise, thereby generating a unified cross-modal context representation $F_{ctx}$:
\begin{equation}
    F_{ctx} = \mathcal{G}(F_V, F_T).
\end{equation}
Here, $F_{ctx}$ serves as a domain-adaptive anchor that encapsulates the high-level semantic consensus between the image and the text description.

\textbf{Bidirectional Calibration via Shared Context.}
With the unified context $F_{ctx}$ established, we perform parallel calibration for both modalities. Specifically, we utilize two symmetric Cross-Attention modules. The original modality features ($F_V$ and $F_T$) serve as the Queries ($Q$) to preserve their intrinsic structural or semantic characteristics, while the shared context $F_{ctx}$ serves as both the Keys ($K$) and Values ($V$) to provide aligned guidance.

For the visual branch, the image features query the shared context to highlight semantically relevant regions:
\begin{equation}
    F'_V = \text{CrossAttn}(Q=F_V, K=F_{ctx}, V=F_{ctx}).
\end{equation}

Simultaneously, for the textual branch, the text embeddings query the same context to focus on medically significant keywords aligned with the visual content:
\begin{equation}
    F'_T = \text{CrossAttn}(Q=F_T, K=F_{ctx}, V=F_{ctx}).
\end{equation}

By utilizing $F_{ctx}$ as a shared reference, the VLCM outputs the calibrated features $F'_V$ and $F'_T$. These features are explicitly aligned within the unified medical semantic space, providing a robust foundation for the subsequent dense prediction tasks.

\subsection{Cost-Aggregating Decoder}
Our decoder adopts the cost aggregation strategy, consistent with the CAT-Seg framework \cite{cho2024cat-seg}. It operates primarily in the pixel-text similarity cost space rather than the high-dimensional feature space, thereby better utilizing the VLM's inherent alignment properties. Specifically, the calibrated visual feature $F'_V$ at the low-resolution $1/32$ scale and the calibrated textual feature $F'_T$ are used to compute the initial Cost Volume $C_{init}$. This matrix is iteratively refined through subsequent Spatial Aggregation and Class Aggregation modules, designed to smooth noise and reinforce inter-class semantic relationships. For generating the final high-resolution output, the decoder employs a multi-stage upsampling process: the aggregated cost feature is sequentially upsampled and integrated with the higher-resolution features, specifically the $1/16$ and $1/8$ scale features obtained from the SSE module. This methodology ensures that both semantic consistency from the $1/32$ cost volume and fine-grained spatial accuracy from the higher-resolution feature maps contribute to the final segmentation mask $M$.

\section{experiments}

\subsection{Datasets and Evaluation Metrics}
To comprehensively evaluate the effectiveness and generalization capability of TGC-Net, we conduct experiments on five challenging medical imaging datasets that span both chest X-ray and abdominal CT modalities.

\noindent\textbf{(1) QaTa-COV19 Dataset.}
QaTa-COV19~\cite{degerli2022osegnet} consists of 9,258 COVID-19 chest X-ray radiographs paired with extended clinical notes. Following the official data split used in LViT~\cite{li2023lvit}, we use 5,716 images for training, 1,429 for validation, and 2,113 for testing.

\noindent\textbf{(2) MosMedData+ Dataset.}
MosMedData+~\cite{li2023lvit, morozov2020mosmeddata, hofmanninger2020automatic} contains 2,729 CT slices depicting lung infections and is used to evaluate model performance on CT-based lesion segmentation. We follow the same data partition protocol as LViT~\cite{li2023lvit}, using 2,183 slices for training and 273 slices each for validation and testing.

\noindent\textbf{(3) MSD-Spleen Dataset.}
MSD-Spleen~\cite{simpson2019large} is an abdominal CT dataset dedicated to spleen segmentation, comprising 32 training volumes and 9 test volumes. For consistency and fair comparison, we adopt the evaluation settings used in CRISP-SAM2~\cite{yu2025crisp}.

\noindent\textbf{(4) WORD Dataset.}
The WORD dataset~\cite{luo2022word} is a large-scale abdominal CT benchmark covering 12 abdominal organs, including 80 training and 20 test volumes. We follow the same experimental setup as CRISP-SAM2~\cite{yu2025crisp}.

\noindent\textbf{(5) AbdomenCT-1k Dataset.}
AbdomenCT-1k~\cite{ma2021abdomenct} consists of 1,000 abdominal CT scans across 5 major organs, with 800 scans used for training and 200 for testing. We adopt the same configurations as CRISP-SAM2~\cite{yu2025crisp} to ensure a fair comparison with prior work.

\noindent\textbf{Evaluation Metrics.}
Model performance is quantitatively evaluated using standard metrics widely adopted in medical image segmentation. For the QaTa-COV19 and MosMedData+ datasets, we report both the mean Dice Similarity Coefficient (mDice) and the mean Intersection-over-Union (mIoU) to comprehensively assess segmentation accuracy. For the MSD-Spleen, WORD, and AbdomenCT-1k abdominal CT datasets, we follow previous works and use the mean Dice Similarity Coefficient (mDice) as the primary evaluation metric.

\subsection{Implementation Details}
Our proposed framework is implemented using PyTorch\cite{paszke2019pytorch}. For image resolution consistency within the dual-branch encoder: the input resolution for the CLIP image encoder is set to $336 \times 336$, while the input resolution for the parallel CNN branch is $768 \times 768$. We set the global batch size to 8 for all experiments. Regarding optimization, the initial learning rate is configured based on the dataset: $2 \times 10^{-5}$ for the QaTa-COV19 dataset, $5 \times 10^{-4}$ for the MosMedData+ dataset, and $2 \times 10^{-4}$ for the three abdominal CT datasets (MSD-Spleen, WORD, and AbdomenCT-1k).We utilize the CLIP-ViT-L model as our base Vision-Language Model. During the training process, the parameters of both the CLIP image encoder and the CLIP text encoder are kept frozen to retain their strong pre-trained representation capabilities.

\subsection{Comparison With State-of-the-Art Methods}

\begin{table}[t]
\centering
\caption{Comparisons of current state-of-the-art methods on the QaTa-COVID19 and MosMedData+ datasets.}
\label{tab:segmentation_comparison_with_text_reordered}
\resizebox{\columnwidth}{!}{
\begin{tabular}{llc|cc|cc}
\hline
Method & Venue & Text & \multicolumn{2}{c|}{QaTa-COVID19} & \multicolumn{2}{c}{MosMedData+} \\
 & & & Dice $\uparrow$ & mIoU $\uparrow$ & Dice $\uparrow$ & mIoU $\uparrow$ \\
\hline
U-Net \cite{ronneberger2015u} & MICCAI'15 & $\times$ & 79.02 & 69.46 & 64.60 & 50.73 \\
U-Net++ \cite{zhou2018unet++} & MICCAI'18 & $\times$ & 79.62 & 70.25 & 71.75 & 58.39 \\
nnUNet  \cite{isensee2021nnu} & Nature'21 & $\times$ & 80.42 & 70.81 & 72.59 & 60.36 \\
Swin-UNet \cite{cao2022swin} & ECCV'22 & $\times$ & 78.07 & 68.34 & 63.29 & 50.19 \\
UCTransNet \cite{wang2022uctransnet} & AAAI'22 & $\times$ & 79.15 & 69.60 & 65.90 & 52.69 \\
\hline
GLoRIA \cite{huang2021gloria} & ICCV'21 & $\checkmark$ & 79.94 & 70.68 & 72.42 & 60.18 \\
ViLT \cite{kim2021vilt} & ICML'21 & $\checkmark$ & 79.63 & 70.12 & 72.36 & 60.15 \\
LAVT \cite{yang2022lavt} & CVPR'22 & $\checkmark$ & 79.28 & 69.89 & 73.29 & 60.41 \\
TGANet \cite{tomar2022tganet} & MICCAI'22 & $\checkmark$ & 79.87 & 70.75 & 71.81 & 59.28 \\
Ariadne's Thread \cite{zhong2023ariadne} & MICCAI'23 & $\checkmark$ & \underline{89.78} & \underline{81.45} & \underline{77.75} & 63.60 \\
LVIT \cite{li2023lvit} &  TMI'23 & $\checkmark$ & 83.66 & 75.11 & 74.57 & 61.33 \\
LGA \cite{hu2024lga} & MICCAI'24 & $\checkmark$ & 84.65 & 76.23 & 75.63 & 62.52 \\
RefSegformer \cite{wu2024toward} &  TIP'24 & $\checkmark$ & 84.09 & 75.48 & 74.98 & 61.70 \\
RecLMIS \cite{huang2024cross} &  TMI'24 & $\checkmark$ & 85.22 & 77.00 & 77.48 & \underline{65.07} \\
ARseg \cite{wang2025towards} & MICCAI'25  & $\checkmark$ & 84.09 & 72.64 & 73.24 & 59.82 \\
\hline
TGC-Net (Ours) & & $\checkmark$ & $\mathbf{90.54}$ & $\mathbf{82.71}$ & $\mathbf{80.94}$ & $\mathbf{68.00}$ \\
\hline
\end{tabular}
}
\end{table}

\begin{table}[t]
\centering
\caption{Segmentation performance (Dice score only) on abdominal CT datasets. $^\star$ means assisted with geometric prompts, and $^{\diamond}$ denotes assisted with anatomical prompt.}
\label{tab:simplified_performance_final}
\resizebox{\columnwidth}{!}{
\begin{tabular}{l|ccc}
\hline
Method  & MSD-Spleen & AbdomenCT-1k & WORD \\
\hline
SAM$^\star$ \cite{kirillov2023segment} & 73.82 & 64.63 & 63.26 \\
SAM2$^\star$ \cite{ravi2024sam} & 83.60 & 83.94 & 76.54 \\
\hline
MedSAM$^\star$ \cite{ma2024segment}  & 76.07 & 69.07 & 72.47 \\
I-MedSAM$^\star$ \cite{wei2024medsam} & 81.37 & 74.71 & 67.60 \\
MedSAM-2$^\star$ \cite{zhu2024medical}  & 86.29 & 86.76 & 77.42 \\
CT-SAM3D$^\star$ \cite{guo2025towards}  & 91.83 & 90.05 & 80.33 \\
\hline
LViT \cite{li2023lvit}  & 79.40 & 75.95 & 73.66 \\
CAT$^{\diamond}$ \cite{huang2024cat}  & 94.26 & 88.38 & 83.53 \\
SegVol$^\star$ \cite{du2024segvol}  & 92.79 & 87.70 & 81.65 \\
ZePT \cite{jiang2024zept}  & 94.03 & 91.34 & 84.26 \\
CRiSP-SAM2 \cite{yu2025crisp}& \underline{95.33} & \underline{92.28} & \underline{85.47} \\
\hline
TGC-Net (Ours) & $\mathbf{95.93}$ & $\mathbf{93.53}$ & $\mathbf{85.93}$ \\
\hline
\end{tabular}
}
\end{table}

\textbf{Performance on Chest Datasets (QaTa-COVID19 and MosMedData+):}  
As presented in Table~\ref{tab:segmentation_comparison_with_text_reordered}, language-guided medical image segmentation models (marked with $\checkmark$) consistently outperform vision-only architectures (marked with $\times$), confirming the advantage of incorporating textual priors. Traditional CNN- and Transformer-based models such as U-Net~\cite{ronneberger2015u}, nnUNet~\cite{isensee2021nnu}, and Swin-UNet~\cite{hatamizadeh2021swin} achieve Dice scores below 81.0\% on QaTa-COVID19, reflecting limited semantic reasoning ability.  
In contrast, text-enhanced frameworks such as GLoRIA~\cite{huang2021gloria} and LAVT~\cite{yang2022lavt} demonstrate notable performance gains, validating the importance of multimodal alignment.       
Our proposed TGC-Net achieves new state-of-the-art (SOTA) performance on QaTa-COVID19, obtaining a Dice score of $90.54\%$ and an mIoU of $82.71\%$. This surpasses the previous best method, Ariadne’s Thread~\cite{zhong2023ariadne}, which reported a Dice of $89.78\%$ and an mIoU of $81.45\%$, by absolute margins of $0.76\%$ and $1.26\%$, respectively.
This remarkable improvement highlights the model’s capability to bridge semantic and structural representations through the proposed Tri-Gap Calibration mechanism. 
On the MosMedData+ dataset, TGC-Net obtains Dice and mIoU scores of $80.94\%$ and $68.00\%$, respectively, surpassing the previous best method, RecLMIS~\cite{huang2024cross}, which achieved $77.48\%$ and 65.07, by absolute gains of $3.46\%$ and $2.93\%$.
These results demonstrate that TGC-Net effectively aligns multimodal cues even under domain-specific noise and limited annotation quality, resulting in superior lesion localization and boundary precision.

\textbf{Performance on Abdominal Datasets (MSD-Spleen, AbdomenCT-1k, and WORD):}  
Results on the abdominal CT datasets (Table~\ref{tab:simplified_performance_final}) further confirm the robustness and generalization of TGC-Net.  
These datasets, especially AbdomenCT-1k and WORD, pose additional challenges due to complex structures and diverse organ morphologies.  
We compare TGC-Net with both foundation models (e.g., MedSAM-2~\cite{zhu2024medical}, CT-SAM3D~\cite{guo2025towards}) and recent LGMIS methods (e.g., ZePT~\cite{jiang2024zept}, CRiSP-SAM2~\cite{yu2025crisp}).  
While foundation models such as CT-SAM3D perform strongly (Dice: $91.83\%$ on MSD-Spleen), our TGC-Net surpasses all existing methods, achieving the best Dice scores of $95.93\%$, $93.53\%$, and $85.93\%$ on MSD-Spleen, AbdomenCT-1k, and WORD, respectively.  
Compared to CRiSP-SAM2, the previous SOTA, TGC-Net improves by $0.60\%$, $1.25\%$, and $0.46\%$ across these datasets, respectively.  
These consistent gains across single and multi-organ segmentation tasks underline the framework’s capacity to generalize to highly heterogeneous medical contexts without task-specific tuning.

\begin{figure*}[t]
    \centering
    \includegraphics[width=0.9\linewidth]{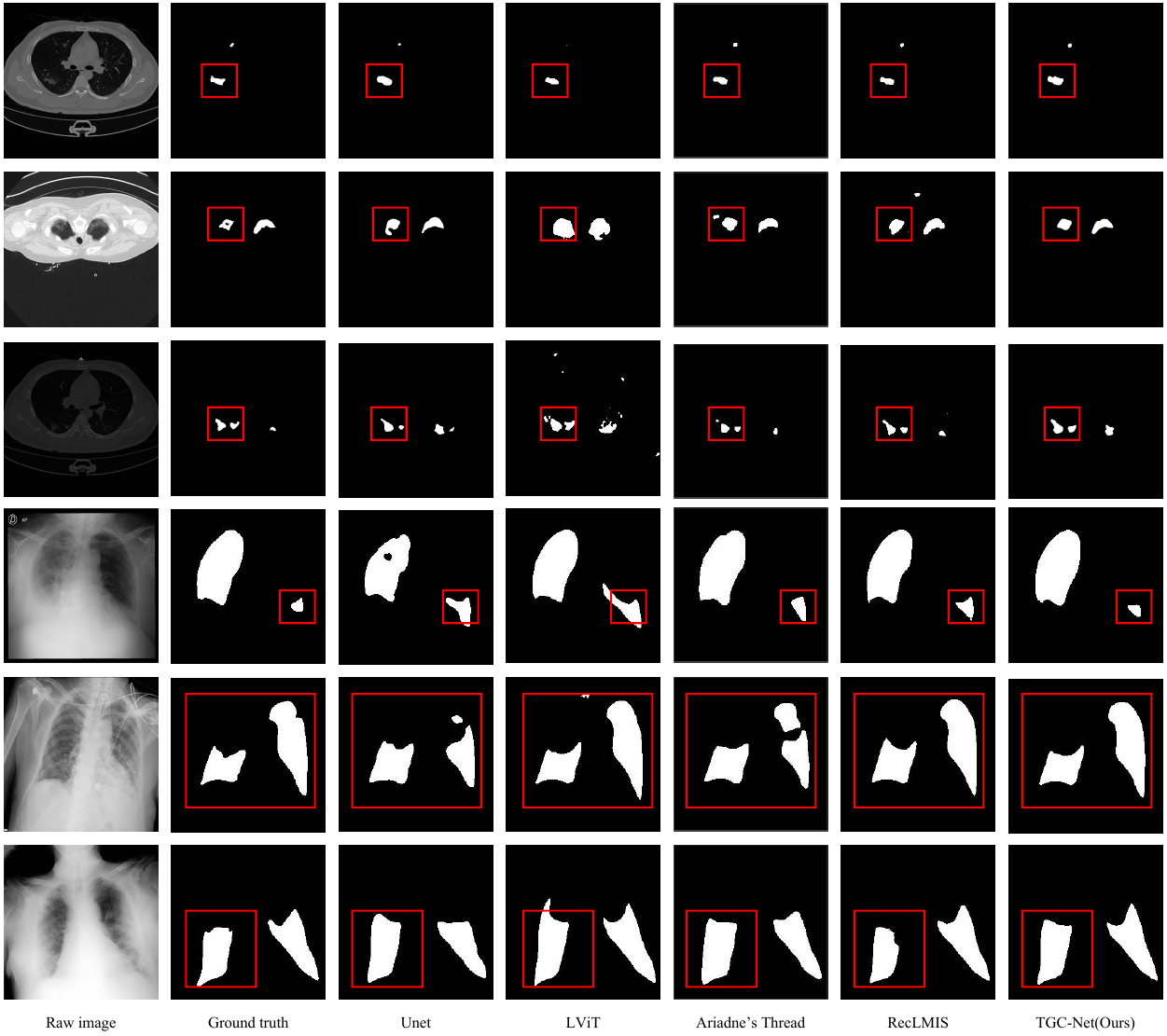}
    \caption{Qualitative comparison on the MosMedData+ and QaTa-COVID19 datasets. The first three rows display results from MosMedData+, while the bottom three rows correspond to QaTa-COVID19.}
    \label{fig:visualation_pdf}
\end{figure*}

\textbf{Qualitative Results:}
Figure~\ref{fig:visualation_pdf} presents representative segmentation results from both datasets, comparing TGC-Net against leading baselines including U-Net, LViT, Ariadne’s Thread, and RecLMIS. Vision-only models such as U-Net often produce coarse or incomplete masks due to their lack of semantic grounding, whereas text-guided approaches (e.g., LViT and Ariadne’s Thread) improve contextual understanding but still exhibit limitations in delineating fine anatomical boundaries.
RecLMIS offers more refined edges but may still generate spurious regions under challenging conditions. In contrast, TGC-Net produces segmentation masks that are both spatially coherent and anatomically faithful to the ground truth. Benefiting from the proposed Tri-Gap Calibration strategy, TGC-Net effectively integrates semantic cues with structural precision, resulting in more accurate localization and clearer boundary preservation.

\textbf{Summary:}  
Across five benchmark datasets spanning X-ray and CT modalities, TGC-Net establishes new SOTA performance and demonstrates strong cross-domain generalization.  
The consistent improvements in both quantitative metrics and qualitative visualization confirm that the proposed framework effectively mitigates the Structural, Semantic, and Alignment Gaps, delivering precise and robust language-guided medical image segmentation.

\subsection{Ablation Study}
\subsubsection{Ablation Study on Core Components}
To validate the effectiveness and necessity of each core component in our TGC-Net, we conduct a comprehensive ablation study. We investigate the individual contributions of the SSE, the DATE, and the VLCM. The experiments are performed on the QaTa-COV19 and MosMedData+ datasets, with results summarized in Table~\ref{tab:ablation}. We use the Dice coefficient (\%) as the primary metric for evaluation.

\begin{table}[htbp] 
\centering
\caption{Ablation study of the core components of TGC-Net on the QaTa-COV19 and MosMedData+ datasets. We report the Dice coefficient (\%). The top row represents our full proposed method.}
\label{tab:ablation}
\resizebox{\columnwidth}{!}{
  \begin{tabular}{c|ccc|cc}
  \toprule
   & \textbf{SSE} & \textbf{DATE} & \textbf{VLCM} & \textbf{QaTa-COV19 (\%)} & \textbf{MosMedData+(\%)} \\
  \midrule
  (a) & \textbf{$\checkmark$} & $\checkmark$ & $\checkmark$ & \textbf{90.54} & \textbf{80.94} \\ 
  \hline
  (b) & $\times$     & $\checkmark$ & $\checkmark$ & 89.55 & 79.38 \\
  (c) & $\checkmark$ & $\checkmark$ & $\times$     & 89.95 & 79.62 \\
  (d) & $\checkmark$ & $\times$      & $\checkmark$ & 89.90 & 79.18 \\ 
  \hline
  (e) & $\times$     & $\times$     & $\times$      & 88.22 & 76.79 \\
  \bottomrule
  \end{tabular}
} 
\end{table}

As shown in Table~\ref{tab:ablation}, our full TGC-Net, with all three modules enabled, achieves the best performance, yielding Dice scores of 90.54\% on QaTa-COV19 and 80.94\% on MosMedData+. We establish a baseline model (last row) by deactivating all three components. This baseline, which relies on a more direct VLM transfer, performs significantly worse, scoring only 88.22\% and 76.79\%, respectively. This performance gap of +2.32\% on QaTa-COV19 and +4.15\% on MosMedData+ underscores the collective importance of our proposed modules in addressing the inherent gaps.

We then analyze the contribution of each module by individually removing it from the full model. 

First, we investigate the effect of SSE. Removing the SSE module (row 2) causes a significant performance drop to 89.55\% ($-0.99\%$) on QaTa-COV19 and 79.38\% ($-1.57\%$) on MosMedData+. This confirms our hypothesis that fusing CNN-based local details with ViT-based global context is critical for bridging the Structural Gap and building a robust visual representation.
    
Next, we examine the effect of DATE. Deactivating the DATE module (row 4) results in scores of 89.90\% ($-0.64\%$) and 79.18\% ($-1.77\%$). This demonstrates the importance of injecting domain-specific medical knowledge into the text encoder to create a semantically rich representation, thereby mitigating the Semantic Gap.

Finally, we assess the effect of VLCM. Removing the VLCM (row 3) leads to a performance of 89.95\% ($-0.73\%$) and 79.62\% ($-1.33\%$). While the SSE and DATE provide strong initial features, this result indicates that an explicit re-alignment step is necessary to bridge the Alignment Gap. The VLCM acts as a crucial calibrator, ensuring the visual and textual features are properly aligned in the medical domain, which provides the final boost to segmentation accuracy.

In summary, the ablation study provides clear evidence that all three proposed components are integral to the success of TGC-Net. Each module effectively addresses its targeted gap, and their synergy is essential for achieving state-of-the-art, language-guided medical image segmentation.

\subsubsection{Ablation Study on the SSE}

To further verify the effectiveness of the proposed SSE and to analyze the contribution of each component, we conducted an ablation study focusing on the interplay between the CNN branch (responsible for local structural cues) and the ViT branch (responsible for global semantic context). Table~\ref{tab:ablation_sse} reports the performance of different encoder configurations on the QaTa-COV19 and MosMedData+ datasets.

\begin{table}[htbp]
\centering
\caption{Ablation study on the SSE. We compare the model performance under different encoder configurations, including CNN-only, ViT-only (CLIP image encoder) with various fine-tuning strategies (frozen backbone, full fine-tuning, and selective fine-tuning of query and value projections), and our final synergistic design (SSE).}
\label{tab:ablation_sse}
\resizebox{\columnwidth}{!}{
\begin{tabular}{l|cc}
\toprule
\textbf{Encoder Configuration} & \textbf{QaTa-COV19} & \textbf{MosMedData+} \\
\midrule
\textbf{TGC-Net w/ SSE (Final)} & \textbf{90.54} & \textbf{80.94} \\
\hline
CNN-only (no CLIP) & 89.17 & 76.81 \\
ViT-only (no fine-tune) & 89.55 & 79.38 \\
ViT-only (qv fine-tune) & 89.60 & 79.86 \\
ViT-only (full fine-tune) & 90.19 & 78.51 \\
CNN + ViT (full fine-tune) & 90.15 & 80.00 \\
\bottomrule
\end{tabular}
}
\end{table}

The results in Table~\ref{tab:ablation_sse} provide several key insights.  

First, the CNN-only baseline (89.17\% / 76.81\%) shows a clear performance degradation compared to our final SSE model. This indicates that relying solely on low-level spatial details fails to capture the high-level semantic priors necessary for robust multimodal alignment. The absence of CLIP-based representations prevents the model from leveraging pretrained vision–language correspondences, thereby limiting its generalization ability.

Second, when using the ViT-only configuration, we observe a moderate performance improvement (up to 89.60\% / 79.86\%) over the CNN-only baseline, confirming the value of pretrained global semantic representations. Interestingly, full fine-tuning of the ViT slightly increases the Dice score on QaTa-COV19 (90.19\%) but causes a drop on MosMedData+ (78.51\%), suggesting that excessive adaptation may distort the pretrained multimodal alignment, especially in smaller or noisier datasets.

Third, we also evaluate a variant that mirrors the SSE architecture but does not freeze the CLIP image encoder. Although this configuration achieves relatively strong results (90.15\% / 80.00\%), it still underperforms the final SSE design. This finding indicates that updating the CLIP backbone encourages the model to overfit the training distribution, thereby weakening the powerful cross-modal priors obtained during large-scale pretraining.

Finally, our SSE-enabled TGC-Net achieves the best overall performance (90.54\% / 80.94\%). This demonstrates that the synergistic interaction between the CNN and ViT branches effectively integrates complementary information: the CNN contributes fine-grained structural refinements, while the ViT preserves high-level semantic coherence. This cooperative mechanism mitigates the Structural Gap and yields a clear “$1 + 1 > 2$” effect, resulting in the most accurate and stable segmentation performance across both datasets.

\subsubsection{Ablation Study on the DATE}

We also conducted experiments to validate our Domain-Augmented Text Encoder (DATE), which addresses the Semantic Gap. Our approach involves injecting external medical knowledge (auxiliary text) into the primary text prompts. We compare this against using only the main text, only the auxiliary text, or text expanded by a general-purpose LLM (e.g., GPT). Results are shown in Table~\ref{tab:ablation_date}.

\begin{table}[htbp]
\centering
\caption{Ablation study on the text encoder (DATE). We compare our auxiliary text injection method against simpler text prompt strategies.}
\label{tab:ablation_date}
\resizebox{\columnwidth}{!}{
\begin{tabular}{l|cc}
\toprule
\textbf{Text Augmentation Method} & \textbf{QaTa-COV19 (\%)} & \textbf{MosMedData+ (\%)} \\
\midrule
\textbf{Auxiliary Text Injection (Ours)} & \textbf{90.54} & \textbf{80.94} \\
\hline
Main Text Only (Baseline) & 89.90 & 79.18 \\
Auxiliary Text Only & 89.71 & 79.02 \\
LLM-Expanded Text & 89.97 & 79.00 \\
\bottomrule
\end{tabular}
}
\end{table}

The ``Main Text Only
baseline (89.90\%, 79.18\%) performs significantly worse than our proposed method, which confirms the findings from our main ablation (Table~\ref{tab:ablation}) that the semantic gap is a real challenge. Interestingly, using ``Auxiliary Text Only" (89.71\%, 79.02\%) or ``LLM-Expanded Text" (89.97\%, 79.00\%) performs even worse. This suggests that the auxiliary text alone lacks specific guidance, and generic LLM expansion may introduce noise or irrelevant information. 

Our DATE method (``Auxiliary Text Injection"), which achieves the top scores (90.54\%, 80.94\%), demonstrates that the optimal strategy is to use the external knowledge to augment and contextualize the primary text prompt, not replace it. This targeted injection of domain knowledge effectively bridges the semantic gap, leading to a more potent textual representation for segmentation guidance.

\subsubsection{Ablation Study on the VLCM}

Finally, we investigate the design choices for our VLCM, which is responsible for bridging the Alignment Gap. We compare our proposed ``Gated Global Alignment Module" against several alternative alignment strategies. The results are presented in Table~\ref{tab:ablation_vlcm}.

\begin{table}[htbp]
\centering
\caption{Ablation study on the design of the VLCM. We compare our final gated global alignment against simpler alignment methods and a baseline with no explicit alignment.}
\label{tab:ablation_vlcm}
\resizebox{\columnwidth}{!}{
\begin{tabular}{l|cc}
\toprule
\textbf{Alignment Method} & \textbf{QaTa-COV19 (\%)} & \textbf{MosMedData+ (\%)} \\
\midrule
\textbf{Gated Global Alignment (Ours} & \textbf{90.54} & \textbf{80.94} \\
\hline
No Prior Alignment (Baseline) & 89.95 & 79.62 \\
Single Cross-Attention & 90.25 & 80.03 \\
Bi-directional Cross-Attention & 90.39 & 80.11 \\
\bottomrule
\end{tabular}
}
\end{table}

We first establish a baseline by removing any explicit alignment module (``No Prior Alignment"). This model (89.95\%, 79.62\%) shows a substantial performance drop of -0.59\% on QaTa-COV19 and -1.32\% on MosMedData+ compared to our final solution. This result strongly confirms that simply providing strong visual and text features is insufficient; an explicit calibration step is essential to resolve the domain-specific Alignment Gap.

Next, we explore standard cross-attention mechanisms as alternatives. Both ``Single Cross-Attention" (90.25\%, 80.03\%) and ``Bi-directional Cross-Attention" (90.39\%, 80.11\%) outperform the baseline, confirming that cross-attention is a viable strategy for fusing and aligning the features. However, our proposed ``Gated Global Alignment (Final VLCM)" (90.54\%, 80.94\%) achieves the highest scores on both datasets. This demonstrates that the sophisticated gated mechanism within our VLCM provides a more effective and nuanced calibration than standard cross-attention, leading to a superior final segmentation.

\subsubsection{Computional Sources}
\begin{figure}[t]
    \centering
    \includegraphics[width=\linewidth]{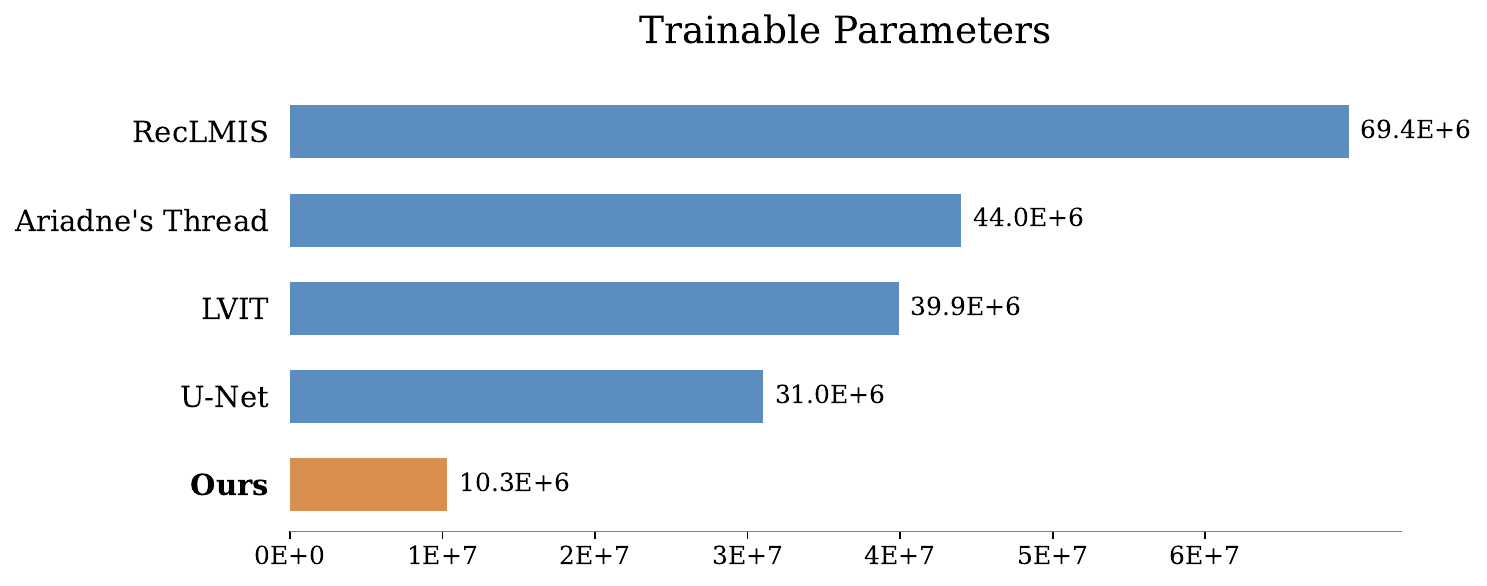}
    \caption{Comparison of trainable parameters across different models.}
    \label{fig:trainable_params}
\end{figure}
To assess the training computational efficiency of TGC-Net, we compare the number of trainable parameters with several representative baselines, including U-Net, LViT, Ariadne’s Thread, and RecLMIS. As shown in Figure~\ref{fig:trainable_params}, TGC-Net contains only 10.3 million trainable parameters, which is substantially fewer than U-Net (31.0M), LViT (39.9M), Ariadne’s Thread (44.0M), and RecLMIS (69.4M).

This efficiency stems from our design choice of freezing all large pretrained components, such as the CLIP vision and text encoders, and training only lightweight modules, including the SSE, DATE, and VLCM, together with the decoder.
By updating only these task-specific components, TGC-Net significantly reduces training cost while preserving the rich priors learned during large-scale pretraining. As a result, our model achieves strong performance with just 10.3M trainable parameters, markedly smaller than many existing approaches.

\section{conclusion}
In this work, we presented TGC-Net, a unified framework designed to adapt the pre-trained CLIP model to the specialized requirements of language-guided medical image segmentation. By addressing the fundamental obstacles of insufficient structural detail, inadequate medical semantics, and misaligned cross-modal representations, we proposed a principled tri-gap calibration strategy incorporating the SSE, DATE, and VLCM. Extensive experiments across five datasets demonstrate that TGC-Net consistently achieves state-of-the-art performance with superior robustness and cross-domain generalization compared to existing approaches. Overall, our framework provides an effective solution for leveraging CLIP in medical imaging, laying a solid foundation for more accurate, reliable, and semantically grounded segmentation.

\section*{Acknowledgements}
This work was funded by the National Natural Science Foundation of China (Grant No. 82472070), the Key Research and Development Program of Hubei Province (Grant No. 2024BCB036), the Hubei Provincial Science and Technology Innovation Talents Program - 2025 Young Science and Technology Talents Training Special Project (Grant No. 2025DJA055) and the Artificial Intelligence Program of Tongji Hospital (AI2024A03).

\begin{thebibliography}{10}
\expandafter\ifx\csname url\endcsname\relax
  \def\url#1{\texttt{#1}}\fi
\expandafter\ifx\csname urlprefix\endcsname\relax\def\urlprefix{URL }\fi
\expandafter\ifx\csname href\endcsname\relax
  \def\href#1#2{#2} \def\path#1{#1}\fi

\bibitem{azad2022medical}
R.~Azad, E.~K. Aghdam, A.~Rauland, Y.~Jia, A.~H. Avval, A.~Bozorgpour, S.~Karimijafarbigloo, J.~P. Cohen, E.~Adeli, D.~Merhof, Medical image segmentation review: The success of u-net. arxiv, arXiv preprint arXiv:2211.14830 (2022).

\bibitem{huang2025multi}
X.~Huang, T.~Zhang, Z.~Xu, J.~Huang, H.~Huang, H.~Yang, B.~Zou, S.~Ding, R.~Ruan, Z.~Huang, et~al., Multi-aspect fusion in foundational large vision model for visible light medical imaging segmentation, Information Fusion (2025) 103385.

\bibitem{liu2025cswin}
X.~Liu, P.~Gao, T.~Yu, F.~Wang, R.-Y. Yuan, Cswin-unet: Transformer unet with cross-shaped windows for medical image segmentation, Information Fusion 113 (2025) 102634.

\bibitem{huang2024cross}
X.~Huang, H.~Li, M.~Cao, L.~Chen, C.~You, D.~An, Cross-modal conditioned reconstruction for language-guided medical image segmentation, IEEE Transactions on Medical Imaging (2024).

\bibitem{li2023lvit}
Z.~Li, Y.~Li, Q.~Li, P.~Wang, D.~Guo, L.~Lu, D.~Jin, Y.~Zhang, Q.~Hong, Lvit: language meets vision transformer in medical image segmentation, IEEE transactions on medical imaging 43~(1) (2023) 96--107.

\bibitem{zhong2023ariadne}
Y.~Zhong, M.~Xu, K.~Liang, K.~Chen, M.~Wu, Ariadne’s thread: Using text prompts to improve segmentation of infected areas from chest x-ray images, in: International Conference on Medical Image Computing and Computer-Assisted Intervention, Springer, 2023, pp. 724--733.

\bibitem{feng2024enhancing}
C.-M. Feng, Enhancing label-efficient medical image segmentation with text-guided diffusion models, in: International Conference on Medical Image Computing and Computer-Assisted Intervention, Springer, 2024, pp. 253--262.

\bibitem{wang2025towards}
Q.~Wang, X.~Lin, Z.~Yan, Towards robust medical image referring segmentation with incomplete textual prompts, in: International Conference on Medical Image Computing and Computer-Assisted Intervention, Springer, 2025, pp. 636--646.

\bibitem{radford2021clip}
A.~Radford, J.~W. Kim, C.~Hallacy, A.~Ramesh, G.~Goh, S.~Agarwal, G.~Sastry, A.~Askell, P.~Mishkin, J.~Clark, et~al., Learning transferable visual models from natural language supervision, in: International conference on machine learning, PmLR, 2021, pp. 8748--8763.

\bibitem{cho2024cat-seg}
S.~Cho, H.~Shin, S.~Hong, A.~Arnab, P.~H. Seo, S.~Kim, Cat-seg: Cost aggregation for open-vocabulary semantic segmentation, in: Proceedings of the IEEE/CVF Conference on Computer Vision and Pattern Recognition, 2024, pp. 4113--4123.

\bibitem{zhang2025perceive}
X.~Zhang, J.~Ma, G.~Wang, Q.~Zhang, H.~Zhang, L.~Zhang, Perceive-ir: Learning to perceive degradation better for all-in-one image restoration, IEEE Transactions on Image Processing (2025).

\bibitem{munia2025attention}
A.~A. Munia, M.~Abdar, M.~Hasan, M.~S. Jalali, B.~Banerjee, A.~Khosravi, I.~Hossain, H.~Fu, A.~F. Frangi, Attention-guided hierarchical fusion u-net for uncertainty-driven medical image segmentation, Information Fusion 115 (2025) 102719.

\bibitem{du2020medical}
G.~Du, X.~Cao, J.~Liang, X.~Chen, Y.~Zhan, Medical image segmentation based on u-net: A review., Journal of Imaging Science \& Technology 64~(2) (2020).

\bibitem{jha2020doubleu}
D.~Jha, M.~A. Riegler, D.~Johansen, P.~Halvorsen, H.~D. Johansen, Doubleu-net: A deep convolutional neural network for medical image segmentation, in: 2020 IEEE 33rd International symposium on computer-based medical systems (CBMS), IEEE, 2020, pp. 558--564.

\bibitem{kayalibay2017cnn}
B.~Kayalibay, G.~Jensen, P.~Van Der~Smagt, Cnn-based segmentation of medical imaging data, arXiv preprint arXiv:1701.03056 (2017).

\bibitem{ronneberger2015u}
O.~Ronneberger, P.~Fischer, T.~Brox, U-net: Convolutional networks for biomedical image segmentation, in: International Conference on Medical image computing and computer-assisted intervention, Springer, 2015, pp. 234--241.

\bibitem{isensee2021nnu}
F.~Isensee, P.~F. Jaeger, S.~A. Kohl, J.~Petersen, K.~H. Maier-Hein, nnu-net: a self-configuring method for deep learning-based biomedical image segmentation, Nature methods 18~(2) (2021) 203--211.

\bibitem{vaswani2017attention}
A.~Vaswani, N.~Shazeer, N.~Parmar, J.~Uszkoreit, L.~Jones, A.~N. Gomez, {\L}.~Kaiser, I.~Polosukhin, Attention is all you need, Advances in neural information processing systems 30 (2017).

\bibitem{chen2021transunet}
J.~Chen, Y.~Lu, Q.~Yu, X.~Luo, E.~Adeli, Y.~Wang, L.~Lu, A.~L. Yuille, Y.~Zhou, Transunet: Transformers make strong encoders for medical image segmentation, arXiv preprint arXiv:2102.04306 (2021).

\bibitem{hatamizadeh2021swin}
A.~Hatamizadeh, V.~Nath, Y.~Tang, D.~Yang, H.~R. Roth, D.~Xu, Swin unetr: Swin transformers for semantic segmentation of brain tumors in mri images, in: International MICCAI brainlesion workshop, Springer, 2021, pp. 272--284.

\bibitem{xia2025comprehensive}
Q.~Xia, H.~Zheng, H.~Zou, D.~Luo, H.~Tang, L.~Li, B.~Jiang, A comprehensive review of deep learning for medical image segmentation, Neurocomputing 613 (2025) 128740.

\bibitem{huang2021gloria}
S.-C. Huang, L.~Shen, M.~P. Lungren, S.~Yeung, Gloria: A multimodal global-local representation learning framework for label-efficient medical image recognition, in: Proceedings of the IEEE/CVF international conference on computer vision, 2021, pp. 3942--3951.

\bibitem{wang2022medclip}
Z.~Wang, Z.~Wu, D.~Agarwal, J.~Sun, Medclip: Contrastive learning from unpaired medical images and text, in: Proceedings of the Conference on Empirical Methods in Natural Language Processing. Conference on Empirical Methods in Natural Language Processing, Vol. 2022, 2022, p. 3876.

\bibitem{huang2024cat}
Z.~Huang, Y.~Jiang, R.~Zhang, S.~Zhang, X.~Zhang, Cat: Coordinating anatomical-textual prompts for multi-organ and tumor segmentation, Advances in Neural Information Processing Systems 37 (2024) 3588--3610.

\bibitem{zhu2020deformable}
X.~Zhu, W.~Su, L.~Lu, B.~Li, X.~Wang, J.~Dai, Deformable detr: Deformable transformers for end-to-end object detection, arXiv preprint arXiv:2010.04159 (2020).

\bibitem{degerli2022osegnet}
A.~Degerli, S.~Kiranyaz, M.~E. Chowdhury, M.~Gabbouj, Osegnet: Operational segmentation network for covid-19 detection using chest x-ray images, in: 2022 IEEE International Conference on Image Processing (ICIP), IEEE, 2022, pp. 2306--2310.

\bibitem{morozov2020mosmeddata}
S.~P. Morozov, A.~E. Andreychenko, N.~A. Pavlov, A.~Vladzymyrskyy, N.~V. Ledikhova, V.~A. Gombolevskiy, I.~A. Blokhin, P.~B. Gelezhe, A.~Gonchar, V.~Y. Chernina, Mosmeddata: Chest ct scans with covid-19 related findings dataset, arXiv preprint arXiv:2005.06465 (2020).

\bibitem{hofmanninger2020automatic}
J.~Hofmanninger, F.~Prayer, J.~Pan, S.~R{\"o}hrich, H.~Prosch, G.~Langs, Automatic lung segmentation in routine imaging is primarily a data diversity problem, not a methodology problem, European radiology experimental 4~(1) (2020) 50.

\bibitem{simpson2019large}
A.~L. Simpson, M.~Antonelli, S.~Bakas, M.~Bilello, K.~Farahani, B.~Van~Ginneken, A.~Kopp-Schneider, B.~A. Landman, G.~Litjens, B.~Menze, et~al., A large annotated medical image dataset for the development and evaluation of segmentation algorithms, arXiv preprint arXiv:1902.09063 (2019).

\bibitem{yu2025crisp}
X.~Yu, C.~Wang, H.~Jin, A.~Elazab, G.~Jia, X.~Wan, C.~Zou, R.~Ge, Crisp-sam2: Sam2 with cross-modal interaction and semantic prompting for multi-organ segmentation, in: Proceedings of the 33rd ACM International Conference on Multimedia, 2025, pp. 1298--1307.

\bibitem{luo2022word}
X.~Luo, W.~Liao, J.~Xiao, J.~Chen, T.~Song, X.~Zhang, K.~Li, D.~N. Metaxas, G.~Wang, S.~Zhang, Word: A large scale dataset, benchmark and clinical applicable study for abdominal organ segmentation from ct image, Medical Image Analysis 82 (2022) 102642.

\bibitem{ma2021abdomenct}
J.~Ma, Y.~Zhang, S.~Gu, C.~Zhu, C.~Ge, Y.~Zhang, X.~An, C.~Wang, Q.~Wang, X.~Liu, et~al., Abdomenct-1k: Is abdominal organ segmentation a solved problem?, IEEE Transactions on Pattern Analysis and Machine Intelligence 44~(10) (2021) 6695--6714.

\bibitem{paszke2019pytorch}
A.~Paszke, S.~Gross, F.~Massa, A.~Lerer, J.~Bradbury, G.~Chanan, T.~Killeen, Z.~Lin, N.~Gimelshein, L.~Antiga, et~al., Pytorch: An imperative style, high-performance deep learning library, Advances in neural information processing systems 32 (2019).

\bibitem{zhou2018unet++}
Z.~Zhou, M.~M. Rahman~Siddiquee, N.~Tajbakhsh, J.~Liang, Unet++: A nested u-net architecture for medical image segmentation, in: International workshop on deep learning in medical image analysis, Springer, 2018, pp. 3--11.

\bibitem{cao2022swin}
H.~Cao, Y.~Wang, J.~Chen, D.~Jiang, X.~Zhang, Q.~Tian, M.~Wang, Swin-unet: Unet-like pure transformer for medical image segmentation, in: European conference on computer vision, Springer, 2022, pp. 205--218.

\bibitem{wang2022uctransnet}
H.~Wang, P.~Cao, J.~Wang, O.~R. Zaiane, Uctransnet: rethinking the skip connections in u-net from a channel-wise perspective with transformer, in: Proceedings of the AAAI conference on artificial intelligence, Vol.~36, 2022, pp. 2441--2449.

\bibitem{kim2021vilt}
W.~Kim, B.~Son, I.~Kim, Vilt: Vision-and-language transformer without convolution or region supervision, in: International conference on machine learning, PMLR, 2021, pp. 5583--5594.

\bibitem{yang2022lavt}
Z.~Yang, J.~Wang, Y.~Tang, K.~Chen, H.~Zhao, P.~H. Torr, Lavt: Language-aware vision transformer for referring image segmentation, in: Proceedings of the IEEE/CVF conference on computer vision and pattern recognition, 2022, pp. 18155--18165.

\bibitem{tomar2022tganet}
N.~K. Tomar, D.~Jha, U.~Bagci, S.~Ali, Tganet: Text-guided attention for improved polyp segmentation, in: International Conference on Medical Image Computing and Computer-Assisted Intervention, Springer, 2022, pp. 151--160.

\bibitem{hu2024lga}
J.~Hu, Y.~Li, H.~Sun, Y.~Song, C.~Zhang, L.~Lin, Y.-W. Chen, Lga: A language guide adapter for advancing the sam model’s capabilities in medical image segmentation, in: International Conference on Medical Image Computing and Computer-Assisted Intervention, Springer, 2024, pp. 610--620.

\bibitem{wu2024toward}
J.~Wu, X.~Li, X.~Li, H.~Ding, Y.~Tong, D.~Tao, Toward robust referring image segmentation, IEEE Transactions on Image Processing 33 (2024) 1782--1794.

\bibitem{kirillov2023segment}
A.~Kirillov, E.~Mintun, N.~Ravi, H.~Mao, C.~Rolland, L.~Gustafson, T.~Xiao, S.~Whitehead, A.~C. Berg, W.-Y. Lo, et~al., Segment anything, in: Proceedings of the IEEE/CVF international conference on computer vision, 2023, pp. 4015--4026.

\bibitem{ravi2024sam}
N.~Ravi, V.~Gabeur, Y.-T. Hu, R.~Hu, C.~Ryali, T.~Ma, H.~Khedr, R.~R{\"a}dle, C.~Rolland, L.~Gustafson, et~al., Sam 2: Segment anything in images and videos, arXiv preprint arXiv:2408.00714 (2024).

\bibitem{ma2024segment}
J.~Ma, Y.~He, F.~Li, L.~Han, C.~You, B.~Wang, Segment anything in medical images, Nature Communications 15~(1) (2024) 654.

\bibitem{wei2024medsam}
X.~Wei, J.~Cao, Y.~Jin, M.~Lu, G.~Wang, S.~Zhang, I-medsam: Implicit medical image segmentation with segment anything, in: European Conference on Computer Vision, Springer, 2024, pp. 90--107.

\bibitem{zhu2024medical}
J.~Zhu, A.~Hamdi, Y.~Qi, Y.~Jin, J.~Wu, Medical sam 2: Segment medical images as video via segment anything model 2, arXiv preprint arXiv:2408.00874 (2024).

\bibitem{guo2025towards}
H.~Guo, J.~Zhang, J.~Huang, T.~C. Mok, D.~Guo, K.~Yan, L.~Lu, D.~Jin, M.~Xu, Towards a comprehensive, efficient and promptable anatomic structure segmentation model using 3d whole-body ct scans, in: Proceedings of the AAAI Conference on Artificial Intelligence, Vol.~39, 2025, pp. 3247--3256.

\bibitem{du2024segvol}
Y.~Du, F.~Bai, T.~Huang, B.~Zhao, Segvol: Universal and interactive volumetric medical image segmentation, Advances in Neural Information Processing Systems 37 (2024) 110746--110783.

\bibitem{jiang2024zept}
Y.~Jiang, Z.~Huang, R.~Zhang, X.~Zhang, S.~Zhang, Zept: Zero-shot pan-tumor segmentation via query-disentangling and self-prompting, in: Proceedings of the IEEE/CVF Conference on Computer Vision and Pattern Recognition, 2024, pp. 11386--11397.

\end{thebibliography}
\end{document}